\documentclass[runningheads]{eccv2020_template/llncs}
\usepackage{graphicx}
\usepackage{tikz}
\usepackage{comment}
\usepackage{amsmath,amssymb} 

\usepackage[colorlinks=true, allcolors=blue]{hyperref}
\usepackage{color}
\usepackage{makecell}
\usepackage{threeparttable}
\usepackage{multicol}
\usepackage{booktabs}
\usepackage{subcaption}
\usepackage[inline]{enumitem}

\usepackage[width=122mm,left=12mm,paperwidth=146mm,height=193mm,top=12mm,paperheight=217mm]{geometry}

\makeatletter
\def\@fnsymbol#1{\ensuremath{\ifcase#1\or *\or \dagger\or \ddagger \or \mathsection \or \mathparagraph\or \|\or **\or \dagger\dagger \or \ddagger\ddagger \else\@ctrerr\fi}}
\makeatother

\makeatletter
\newcommand{\printfnsymbol}[1]{%
  \textsuperscript{\@fnsymbol{#1}}%
}
\makeatother

\begin{document}
\pagestyle{headings}
\mainmatter
\def\ECCVSubNumber{30}

\title{DMD: A Large-Scale Multi-Modal \\ Driver Monitoring Dataset for \\ Attention and Alertness Analysis}

\titlerunning{DMD:~A~Driver~Monitoring~Dataset for~Attention~and~Alertness~Analysis}

\author{
{Juan Diego}~Ortega\printfnsymbol{1}\inst{1,3} %
\and %
Neslihan~Kose\printfnsymbol{1}\inst{2} %
\and %
Paola~Ca\~{n}as\printfnsymbol{2}\inst{1} %
\and %
Min-An~Chao\printfnsymbol{2}\inst{2} %
\and %
Alexander~Unnervik\inst{2} %
\and %
Marcos~Nieto\inst{1} %
\and %
Oihana~Otaegui\inst{1} %
\and %
Luis~Salgado\inst{3} %
}

\authorrunning{J.D. Ortega et al.}

\institute{
Vicomtech~Foundation,~Basque~Research~and~Technology~Alliance~(BRTA),~Spain \\
\and
Dependability~Research~Lab,~Intel~Labs~Europe,~Intel~Deutschland~GmbH,~Germany \\
\and
ETS~Ingenieros~de~Telecomunicaci\'on,~Universidad~Polit\'ecnica~de~Madrid,~Spain \\
\url{https://dmd.vicomtech.org/}
}
\maketitle

\newcounter{mycounter}
\renewcommand{\thefootnote}{\fnsymbol{mycounter}}
\footnotetext[1]{\makeatletter\setcounter{footnote}{0}\renewcommand{\thefootnote}{\fnsymbol{footnote}}{\footnotemark[1]},{\footnotemark[2]}\makeatother~Means equal contribution, respectively.}
\setcounter{footnote}{0}\renewcommand{\thefootnote}{\arabic{footnote}}

\begin{abstract}
Vision is the richest and most cost-effective technology for Driver Monitoring Systems (DMS), especially after the recent success of Deep Learning (DL) methods. The lack of sufficiently large and comprehensive datasets is currently a bottleneck for the progress of DMS development, crucial for the transition of automated driving from SAE Level-2 to SAE Level-3. In this paper, we introduce the \textit{Driver Monitoring Dataset (DMD)}, an extensive dataset which includes real and simulated driving scenarios: distraction, gaze allocation, drowsiness, hands-wheel interaction and context data, in 41 hours of RGB, depth and IR videos from 3 cameras capturing face, body and hands of 37 drivers. A comparison with existing similar datasets is included, which shows the DMD is more extensive, diverse, and multi-purpose.
The usage of the DMD is illustrated by extracting a subset of it, the \textit{dBehaviourMD} dataset, containing 13 distraction activities, prepared to be used in DL training processes. Furthermore, we propose a robust and real-time driver behaviour recognition system targeting a real-world application that can run on cost-efficient CPU-only platforms, based on the \textit{dBehaviourMD}. Its performance is evaluated with different types of fusion strategies, which all reach enhanced accuracy still providing real-time response.  

\end{abstract}

\section{Introduction}
\label{sec:intro}
Road accidents affect drivers and passengers, but also vulnerable pedestrians, motorcyclists and cyclists. Human factors are a contributing cause in almost 90\% of those road accidents \cite{Dingus2016}. Therefore, there has been an active support to the development of fully automated vehicles whose aim is to reduce crashes due to driver factors, eventually achieving the desired EU \textit{Vision Zero objective} road scenario \cite{EuropeanCommission2011}. As automated driving technology advances through SAE-L2 and SAE-L3 \cite{SAE2018}, there will be a paradigm shift in terms of driver responsibility and function. The driving task will become a shared activity between the human and the machine \cite{Fridman2018}. For this to be successful, modern driving automation systems shall increasingly include DMS to measure features of the driver inside the cabin and determine the driver's readiness to perform the driving task. Such systems are specially valuable to guarantee a more reliable and safer mode transfer to operate the vehicle \cite{Mioch2017}.    

DMS have been developed for the past decade as a measure to increase road safety and also to improve driving comfort \cite{Sikander2019,McDonald2019}. The objective of such systems is to understand driver's state according to measurable cues extracted either from direct observation of the driver (i.e. physiological parameters \cite{Chowdhury2018}, visual appearance of the driver \cite{Fernandez2016}), indirect activity of on-board sensors (i.e. vehicle dynamics, cellphone, smart watches or GPS) \cite{Aghaei2016} or a combination of them \cite{JacobedeNaurois2019}. Computer vision methods have acquired important attention particularly due to the demonstrated capabilities of the emerging DL technologies \cite{Rasouli2018}. These methods have a good balance between robustness and applicability, since they could achieve high accuracy rates with non-obtrusive sensors~\cite{Fridman2018}. However, these techniques demand large amounts of visual data to build the necessary models to achieve the desired task (i.e. detect distraction, drowsiness, involvement in secondary tasks, etc.)~\cite{Abraham2017}. There is a significant lack of comprehensive DMS-related datasets open for the scientific community, which is slowing down the progress in this field, while, in comparison, dozens of large datasets exist with sensing data for the exterior of the vehicle~\cite{Janai2017}.

In this paper, we introduce a large-scale multi-modal Driver Monitoring Dataset (DMD) which can be used to train and evaluate DL models to estimate alertness and attention levels of drivers in a wide variety of driving scenarios. The DMD includes rich scene features such as varying illumination conditions, diverse subject characteristics, self-occlusions and 3 views with 3 channels~(RGB/Depth/IR) of the same scene. With the DMD, we want to push forward the research on DMS, making it open and available for the scientific community. In addition, we show how to extract a subset of it (the \textit{dBehaviourMD}) and prepare it as a training set to train and validate a DL method for L3 AD by providing a real-time, robust and reliable driver action detection solution.

\section{Review of in-vehicle monitoring datasets and methods}
\label{sec:soa}
Driver monitoring can be interpreted as observing driver's features related to distraction/inattention, the direction of gaze, head pose, fatigue/drowsiness analysis, disposition of the hands, etc. The state-of-the art shows approaches for these aspects individually and also in a holistic way. Existing literature can be categorised according to the driver's body-part of interest showing visible evidence of different types or aspects of the driver's behaviour.

\textbf{Hands-focused approaches:}
The position of hands and their actions have shown a good correlation to the driver's ability to drive. The work carried out in this matter includes the creation of datasets focused only on hands, like CVRR-HANDS 3D \cite{Ohn-Bar2013} or the VIVA-Hands Dataset \cite{Das2015}, which include the annotation of hands' position in images as bounding boxes. Example methods that use these datasets detect the position of hands using Multiple Scale Region-based Fully Convolutional Networks (MS-RFCN) \cite{Le2017} and detection of interaction with objects such as cellphones to estimate situation which can distract the driver \cite{Le2016}. 

\textbf{Face-focused approaches:}
The driver's condition in terms of distraction, fatigue, mental workload and emotions can be derived from observations of the driver's face and head \cite{Sikander2019}. One aspect widely studied is the direction of eye gaze as in DR(eye)VE \cite{Palazzi2018}, DADA \cite{fang2019dada} and BDD-A \cite{Xia2017} datasets. These features plus information about the interior of the vehicle allows identifying which specific areas of the cabin are receiving the attention of the driver, like in DrivFace \cite{Diaz-Chito2016}, with the possibility of generating attention heat maps or buffer metrics of distraction such as AttenD \cite{Ahlstrom2013b}. However, many of the available datasets are not obtained in a driving environment, normally they are captured in a laboratory with a simulation setup such as Columbia dataset~\cite{CAVE_0324} and MPIIGaze dataset~\cite{xucongZhang2017}. Moreover, datasets such as DriveAHead~\cite{schwarzHaurilet2017} and DD-Pose\cite{Roth2019} have appeared consisting of images of the driver's head area and are intended to estimate the head pose, containing annotations of yaw, pitch and roll angles.   

Fatigue is an important factor that does not allow the driver to safely perform the driving task. Computer vision methods that tackle the detection of fatigued drivers rely on the extraction of parameters from head, mouth and eye activity~\cite{Ortega2020}. In this context, datasets for the identification of drowsy drivers vary in the type of data they offer. They may contain images with face landmarks like DROZY~\cite{massozLangohr2016} and NTHU-DDD~\cite{Weng2017}. With these facial points, more complex indicators can be extracted such as head position, 3D head pose and face activity~\cite{Goenetxea2018}, blink duration~\cite{Baccour2019}, frequency and PERCLOS~\cite{Trutschel2011} with the final goal of estimating the level of drowsiness of the driver~\cite{Mandal2017}.

\textbf{Body-focused approaches:} A side-view camera can further extend the effective monitoring range to the driver’s body action. The first large-scale image dataset for this purpose is the StateFarm’s dataset \cite{StateFarm2016}, which contains 9 distracted behaviour classes apart from safe driving. However, its use is limited to the purposes of the competition. A similar image-based open dataset is the AUC Distracted Driver (AUC DD) dataset \cite{Abouelnaga2018}. In~\cite{Abouelnaga2018}, driver behaviour monitoring with a side-view camera was approached by image-based models, which cannot capture motion information. By approaching it as video-based action recognition problem, the image-based AUC Dataset \cite{Abouelnaga2018} is adapted for spatio-temporal models in \cite{kose19_itsc} which proves that just by incorporating the temporal information, significant increase can be achieved in classification accuracy.

The analysis of RGB-D images has become common practice to estimate the position of the body in a 3D space \cite{CrayeK15}. The Pandora Dataset \cite{borghi2017poseidon} provides high resolution RGB-D images to estimate the driver's head and shoulders' pose. Works like \cite{deoTrivedi2018}, whose objective is to identify driver’s readiness, study various perspectives of the body parts (face, hands, body and feet). More recently, the dataset Drive\&Act \cite{martin2019_iccv} was published containing videos imaging the driver with 5 NIR cameras in different perspectives and 3 channels (RGB, depth, IR) from a side view camera; the material shows participants performing distraction-related activities in an automated driving scenario.

In comparison with our DMD, most of the datasets found in the literature focus either on specific parts of the drivers' body or specific driving actions. With the DMD dataset, we provide a wider range of activities which are directly related to the driver's behaviour and attention state.
Table~\ref{tab:comp_monitor_db} shows a detailed comparison of the studied dataset's features. As can be seen, none of the existing datasets contain DMD's variability, label richness and volume of data.

The closest dataset to DMD in terms of volume and label richness is the recent Drive\&Act \cite{martin2019_iccv}. However, there is major difference, as Drive\&Act focuses on automated driving actions, where the driver is completely not engaged in the driving task, the participants can perform more diverse actions (e.g. working-on-laptop). In our understanding, this dataset is of interest for future SAE L4-5 automated cars, while DMD focuses on a wider domain of driving behaviours (not only actions) which are still non-solved challenges for the automotive industry for SAE L2-3 AD. 

\begin{table}[t]
	\caption{Comparison of public vision-based driver monitoring datasets.}
	\label{tab:comp_monitor_db}
	\centering
	\resizebox{\textwidth}{!}{%
	\begin{threeparttable}
	\begin{tabular}{lcccrccccc}
		\toprule
		{Dataset}    & {Year} & {Drivers\tnote{a}} & {Views\tnote{b}} & {Size\tnote{c}} & {GT\tnote{d}} & {Streams} & {Occlusions} & {Scenarios} & {Usage} \\
		\midrule
		\midrule
		CVRR-Hands\cite{Ohn-Bar2013} & 2013 & 8~(1/7) & 1 & 7k & \makecell{Hands,\\ Actions} & \makecell{RGB\\ Depth} & Yes & Car & \makecell{Normal driving,\\ Distraction} \\  
		\midrule
		{DrivFace\cite{Diaz-Chito2016}} & 2016 & 4~(2/2) & 1 & 0.6k & \makecell{Face/Head} & {RGB} & No & Car & \makecell{Normal driving,\\ Head pose} \\
		\midrule
		{DROZY\cite{Massoz2016}} & 2016 & 14~(11/3) & 1 & 7h & \makecell{Face/Head\\Physiological} & {IR} & No & Laboratory & \makecell{Drowsiness} \\
		\midrule
		{NTHU-DDD\cite{Weng2017}} & 2017 & 36~(18/18) & 1 & 210k & \makecell{Actions} & \makecell{RGB\\ IR} & Yes & Simulator & \makecell{Normal driving,\\ Drowsiness} \\
		\midrule
		{Pandora\cite{borghi2017poseidon}} & 2017 & 22~(10/12) & 1 & 250k & \makecell{Face/Head, \\ Body} & \makecell{RGB\\ Depth} & Yes & Simulator & \makecell{Head/Body pose} \\ 
		\midrule
		{DriveAHead\cite{schwarzHaurilet2017}} & 2017 & 20~(4/16) & 1 & 10.5h & \makecell{Face/Head, \\ Objects} & \makecell{Depth\\ IR} & Yes & Car & \makecell{Normal driving,\\ Head/Body pose} \\ 
		\midrule
		{DD-Pose\cite{Roth2019}} & 2019 & 24~(6/21) & 2 & 6h & \makecell{Face/Head, \\ Objects} & \makecell{RGB\tnote{e}\\ Depth\tnote{f}\\ IR\tnote{f}} & Yes & Car & \makecell{Normal driving,\\ Head/Body pose} \\
		\midrule
		{AUC-DD\cite{Eraqi2019}} & 2019 & 44~(15/29) & 1 & 144k & Actions & {RGB} & No & Car & \makecell{Normal driving,\\ Distraction} \\
		\midrule
		{Drive\&Act\cite{martin2019_iccv}} & 2019 & 15~(4/11) & 6 & 12h & \makecell{Hands/Body,\\ Actions,\\ Objects} & \makecell{RGB\tnote{e}\\ Depth\tnote{e}\\ IR} & No & Car & \makecell{Autonomous driving,\\ Distraction} \\
		
		\midrule
		\midrule
		\textbf{DMD} & 2020 & 37 (10/27) &  3 & 41h & \makecell{Face/Head,\\ Eyes/Gaze,\\ Hands/Body,\\ Actions,\\ Objects} & \makecell{RGB\\ Depth\\ IR} & Yes & \makecell{Car,\\ Simulator} & \makecell{Normal driving,\\ Distraction,\\ Drowsiness} \\
		\bottomrule
	\end{tabular}
	\begin{tablenotes}[flushleft]
	    \setlength{\columnsep}{0.5cm}
        \setlength{\multicolsep}{0cm}
	    \begin{multicols}{3}
    	    \item[a]Number of drivers~(female/male)
    	    \item[b]Simultaneous views of scene 
    	    \item[c]h: hours of video, k: image number 
    	    \item[d]Ground-truth data
    	    \item[e]Only for side view
    	    \item[f]Only for face view
    	\end{multicols}
	\end{tablenotes}
	\end{threeparttable}%
	}
\end{table}

\section{The Driver Monitoring Dataset (DMD)}
\label{sec:dmd}
The DMD is a multi-modal video dataset with multimedia material of different driver monitoring scenarios from 3 camera views. In this paper, one example use-case of the dataset is shown for a real-world application. We propose a robust and real-time driver behaviour recognition system using the multi-modal data available in DMD with a cost-efficient CPU-only platform as computing resource.

\subsection{Dataset specifications}
\label{subsec:specs}
The DMD was devised to address different scenarios in which driver monitoring must be essential in the context of automated vehicles SAE L2-3. 

\begin{figure}[t]
    \centering
    \includegraphics[width=0.95\textwidth]{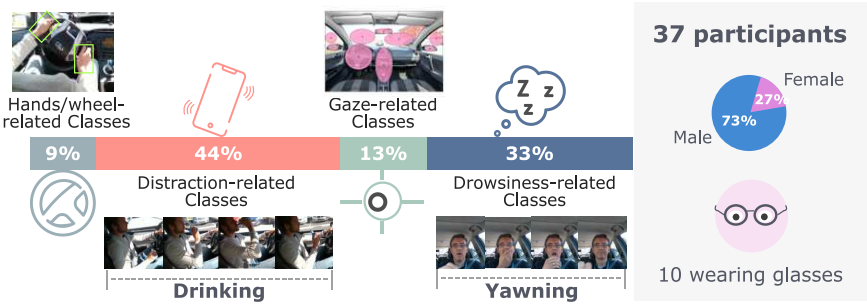}
    \caption{Class distribution of the complete DMD dataset. Both geometrical and temporal-based annotation are included in the available DMD ground truth data.}
    \label{fig:classes}
\end{figure}

The dataset is composed of videos of drivers performing \textit{distraction} actions, drivers in different states of \textit{fatigue and drowsiness}, specific material for \textit{gaze allocation} to interior regions, \textit{head-pose estimation} and different driver's \textit{hands' positions and interaction} with inside objects (i.e. steering wheel, bottle, cellphone, etc.). There was a participation of 37 volunteers for this experiment, the gender proportions are 73\% and 27\% for men and women, respectively, 10 wearing glasses~(see Fig.~\ref{fig:classes}). The age distribution of the participants was homogeneous in the range of 18 to 50 years. The participants were selected to assure novice and expert drivers were included in the recordings. Each participant signed an GDPR informed consent which allows the dataset to be publicly available for research purposes. Moreover, for certain groups of participants, the recording sessions were repeated, one day in the morning and another in the afternoon, to have variation in lighting. Recording sessions were carried out in different days to guarantee variety of weather conditions.  

The DMD dataset was designed specifically to cover the lack of data for a fully operational DMS. Different scripted protocols were defined for each of the behaviour domains. However, no further instructions were given about how to perform the specific actions. Therefore, personal variability is introduced by each of the participants, contributing with a realistic component to the available data. In the \textbf{distraction recordings}, the drivers performed the most relevant actions which affects attention allocation namely: mobile phone use, use of infotainment, drinking, combing hair, reaching for an object, change gear and talking to the passenger. Regarding \textbf{drowsiness recordings}, drivers were asked to perform actions comprised by the most correlated signs of fatigue such as reduce eyelid aperture, microsleeps (i.e. small intervals of closed eyes), yawning and nodding. Moreover, during the \textbf{gaze-related recordings} the participants fixated their gaze to 9 predefined regions which covers all the surroundings of the vehicle. Similarly, for the \textbf{hand-wheel interaction recordings} the drivers held the steering wheel with different hands positions. 

Annotation was done according to the recording type and could include: geometrical features (i.e. landmarks points and bounding boxes), temporal features (i.e. events and actions) or context. Each driver behaviour type has a diverse set of both geometric and temporal classes. Fig.~\ref{fig:classes} depicts the full distribution of classes available in the DMD. This distribution for the different types of recordings was done using a total of 93 classes (temporal, geometric and context). In this paper we focus on a subset of the distraction scenario in which we annotate temporal actions used by the DL algorithm to identify driver actions. The open annotation format VCD\footnote{https://vcd.vicomtech.org/} was chosen to describe the sequences. 

\subsection{Video streams}
\label{subsec:streams}
The dataset was recorded with Intel Realsense D400 Series Depth Cameras which can capture Depth information, RGB and Infrared (IR) images synchronously. Two environments was considered:
\begin{enumerate*}[label=(\roman*)]
    \item outdoors in a real car and
    \item indoors in a driving simulator that immerses the person to close-to-real driving conditions.
\end{enumerate*}
The simulator allows recording the actions or states that are riskier to perform in the car and to obtain the drivers’ natural reactions. Such actions were also recorded in the car while being stopped.

\begin{figure}[t]
    \centering
    \includegraphics[width=\textwidth]{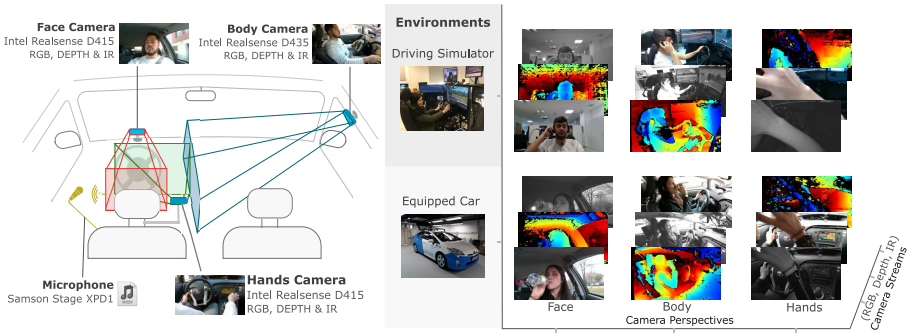}
    \caption{DMD camera setup for both real car and simulator scenarios.}
    \label{fig:setup}
\end{figure}

The recordings were obtained with 3 cameras placed in the aforementioned environments (see Fig.~\ref{fig:setup}). Two D415 cameras recorded the face area frontally and the hands area from the back; one D435 camera was used to capture the driver’s body from the left side. The three cameras capture images at around 30 FPS. After a post-processing step, all camera streams were synchronised to a common frame rate. The resulting material consists of mp4 video files for each stream from the 3 cameras (9 mp4 files for each scene). The raw material reaches a volume of 26 TB (source ROS bags), while the mp4 files are slightly compressed with H.264, with target bitrate 15000 kb/s to ease downloading and processing.

In this paper, we also present a driver monitoring function using the data collected with the side-view camera and the sequences of the driving actions scenario only. For this purpose, initially, the side view data is further processed to short video clips, which enables real-time response from the monitoring system, in order to establish a larger scale multi-modal driver behaviour monitoring dataset named as \textit{dBehaviourMD}\footnote{Both \textit{DMD} and \textit{dBehaviourMD} will be publicly available for the research community.}.

\subsection{Driver behaviour monitoring dataset (dBehaviourMD)}

The side-view recording is done with Intel RealSense D435, which provides RGB frames up to $1920\times1080$ pixels, and IR and depth frames up to $1280\times720$ pixels. The FoV of IR and depth are the same, and their pixels are aligned. However, since RGB camera has slightly narrower FoV and different camera intrinsics than IR, pixel-wise alignment of RGB frames with IR and depth frames requires the camera intrinsic parameters after calibration. Although the intrinsic parameters of the cameras are available from the hardware, in this paper, RGB frames are used directly without calibration and rectification for the experiments. 

For efficiency purposes, all the frames are resized to $640\times360$ pixels for the \textit{dBehaviourMD}. Each entry in a depth array $d_m$ is measured in meters, and is converted to pixel value $d_p \in [0, 255]$ by:
\begin{eqnarray}
d_p = \left\{ 
\begin{array}{ll}
255 \cdot \min(d_{\text{min}} / d_m, 1)
& \text{ if } d_m > 0 \\
0 & \text{ otherwise }
\end{array}
\right. ,
\end{eqnarray}
where $d_{\text{min}} = 0.5$ meter, meaning $d_m < d_{\text{min}}$ will be saturated to 255, otherwise the closer the brighter.

For real-time recognition, the machine has to monitor the driver through a sliding window with a fixed time span to recognise the action and give an instant response. The time span should be as short as possible, so that the machine can be trained to react faster.
In this paper, the time span is set to 50 frames (1.67 seconds) according to our observations on the actions in our dataset. Shorter clips are preserved as the action can be easily recognised during annotation. Longer clips are split into short clips so that the average duration of clips approaches this target length.

\begin{figure}[t!]
    \begin{subfigure}[t]{0.5\textwidth}
        \centering
        \includegraphics[height=4.2cm]{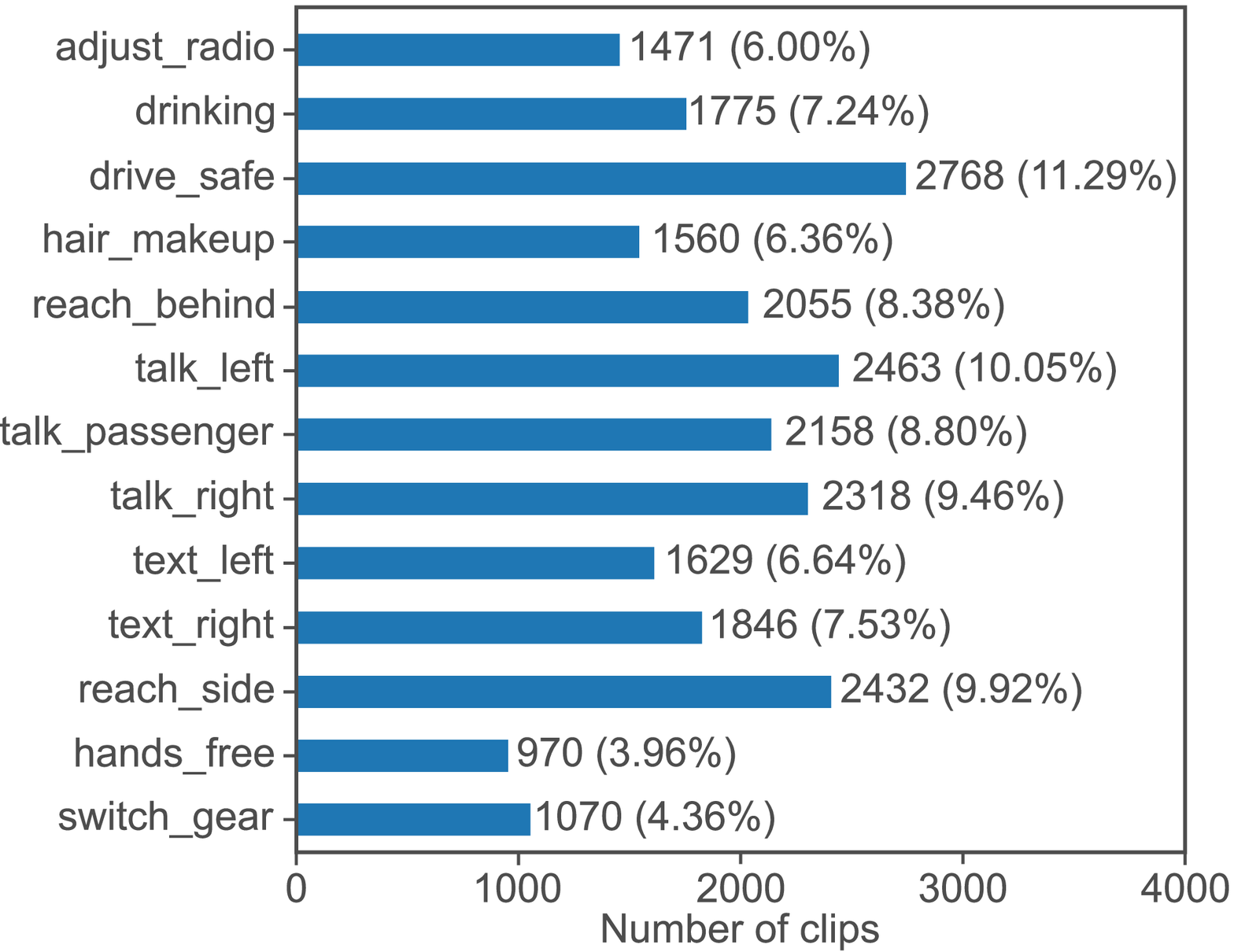}
        \caption{}
        \label{fig:sub_clip_dist}
    \end{subfigure}
    \begin{subfigure}[t]{0.5\textwidth}
        \centering
        \includegraphics[height=4.2cm]{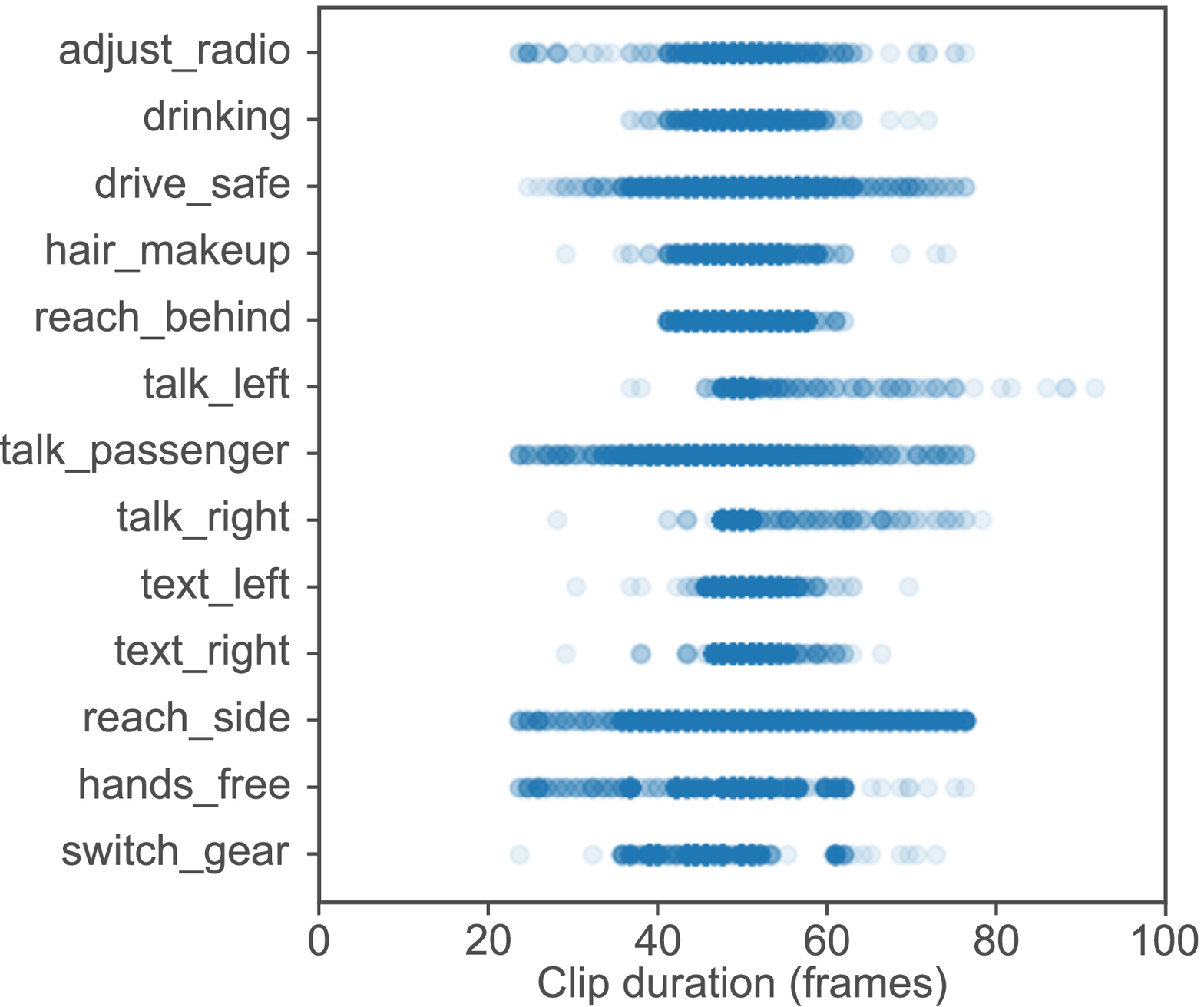}
        \caption{}
        \label{fig:sub_clip_duration}
    \end{subfigure}
	\caption{\textit{dBehaviourMD} class distribution (a) and clip duration distribution for each class (b) after clip splitting and balance sampling.}
	\label{fig:vidas_frames_orig_bal}
\end{figure}

Next, sampling operations\footnote{down/upsampling according to the number of samples in each action class.} have been applied to have a balanced dataset, which has the class distribution and the clip duration distribution for each class
as shown in Fig.~\ref{fig:vidas_frames_orig_bal} (a) and Fig.~\ref{fig:vidas_frames_orig_bal} (b), respectively. It can be noticed that although the average clip duration is similar, there are some classes with relatively various duration than others, meaning those actions are temporally more diverse. In \textit{dBehaviourMD}, the standard deviation of clip duration is kept relatively small,
resulting in a dataset that is suitable for deployment with a fixed timing window. Finally, the training, validation, and testing clip sets are split with the ratio roughly 4:1:1, which corresponds to 24:7:6 drivers, respectively.

The comparison of dBehaviourMD with existing datasets for side-view monitoring is shown in Table~\ref{tab:ds_comp_vidas_dbm}. 
dBehaviourMD has data collected from more number of drivers in a real-driving scenario compared to existing datasets and it has the 10 classes of the StateFarm \cite{StateFarm2016} and AUC DD \cite{Abouelnaga2018} datasets, which are shown in Fig.~ \ref{fig:vidas_frames_orig_bal}, plus 3 classes \textit{reaching side}, \textit{hands free} and \textit{switch gear} which appear quite often during real driving. 
The most recent multi-modal video dataset is Drive\&Act \cite{martin2019_iccv}, which has overall 83 classes annotated in a multi-hierarchical way, being only 34 of them semantically related to driving behaviour or driver action (initially, 12 high-level instructions are ordered to the drivers. Each of them consists of a fixed routine of several second-level actions. Next, 34 fine-grained activities are defined, including for example sitting still, drinking, etc.). 

\begin{table}[tbp]
	\caption{Specifications of dBehaviourMD and existing driver behaviour datasets.}
	\label{tab:ds_comp_vidas_dbm}
	\centering
	\resizebox{0.8\textwidth}{!}{
	\begin{tabular}{lccc}
		\hline
		{}   & {AUC DD~\cite{Abouelnaga2018}} & {Drive\&Act~\cite{martin2019_iccv}} & \textbf{dBehaviourMD} \\
		\hline
		\hline
		Target actions               & Driving behaviour & General behaviour  & Driving behaviour \\ 
		Real driving                 & No               & No                & Yes \\
		Video-based                  & No               & Yes               & Yes \\
		\hline                                           
		\# Classes  & 10 & 34 (Activities)      & 13 \\
		Avg. video clips    & N/A & 303           & 1886 \\
		per class (Min./Max.)                &                  & (19/2797)         & (970/2768) \\
		\hline
	\end{tabular}}
\end{table}

Being a multi-modal video dataset, which enables to capture temporal information efficiently, and having the most number of average video clips per class, the introduced dBehaviourMD dataset outperforms the existing datasets recorded for behaviour recognition purposes.    

\section{Real-time driver behaviour recognition}
\label{sec:usecase}

To develop a real-time system which monitor driver behaviour robustly and reliably, several requirements have to be fulfilled:
\begin{itemize}
\item a cost-effective CNN model running on video data, which provides higher accuracy benefiting from both spatial and temporal information;
\item a video dataset recorded in a real-driving scenario which not only has RGB modality but also infrared (IR) and depth modalities to benefit from complementary information of different modalities; and
\item an architecture which can accommodate more than one modality and fuse them on-the-fly to target a real-time deployment.
\end{itemize}

Behaviour recognition is achieved via spatio-temporal analysis on video data that could be done in several ways such as applying 2D CNNs on multiple video frames or 3D CNNs on video clips. Temporal segment network (TSN) \cite{wang16_eccv,wang19_tpami} applies 2D CNNs for action recognition with sparse temporal sampling and uses optical flow for the temporal stream. Motion fused frames (MFF) \cite{kopu18_cvprw} proposes a new data-level fusion strategy and has more capabilities to learn spatial-temporal features with less computational effort in the base model. Optical flow dimensions in MFF contain the frame-wise short-term movement, while long-term movement is kept across the segments. Both TSN and MFF use optical flow which is not very efficient for real-time deployment.  

Apart from 2D-CNN models, 3D-CNNs have also been applied for action recognition \cite{tran15_iccv,carr17_cvpr,xie18_eccv,tran18_cvpr,kopu19}. 2D-CNN models usually do not take all the consecutive frames, and recent studies in 3D-CNN also question the necessity of 3D-CNN \cite{xie18_eccv}
and consecutive frames for general action recognition \cite{tran18_cvpr}. For a fair comparison, the authors of \cite{tran18_cvpr} compare the models with the identical ResNet building blocks which differ only in the temporal dimension, and they observe that the temporal information contributes not as much as the spatial one. \cite{tran15_iccv} examines the accuracy-speed trade-off claiming the 3D model can bring 3\% accuracy gain with the double computational effort compared with the 2D model on the mini-Kinetics-200 dataset. Although 3D-CNN models could extract more temporal information than 2D-CNN ones, the overhead computational cost is also significant, making it difficult for real-time deployment. 

According to \cite{hara18_cvpr}, the datasets used for transfer learning contribute more than the architectural difference and hence 2D-CNN models pre-trained on ImageNet \cite{deng09_cvpr} can still perform better or at least similarly than 3D-CNN models pre-trained on Kinetics~\cite{carr17_cvpr}.  

\subsubsection{Proposed architecture for real-time deployment:}

\begin{figure}[t!]
    \begin{subfigure}[t]{0.55\linewidth}
        \centering
        \includegraphics[height=3.5cm]{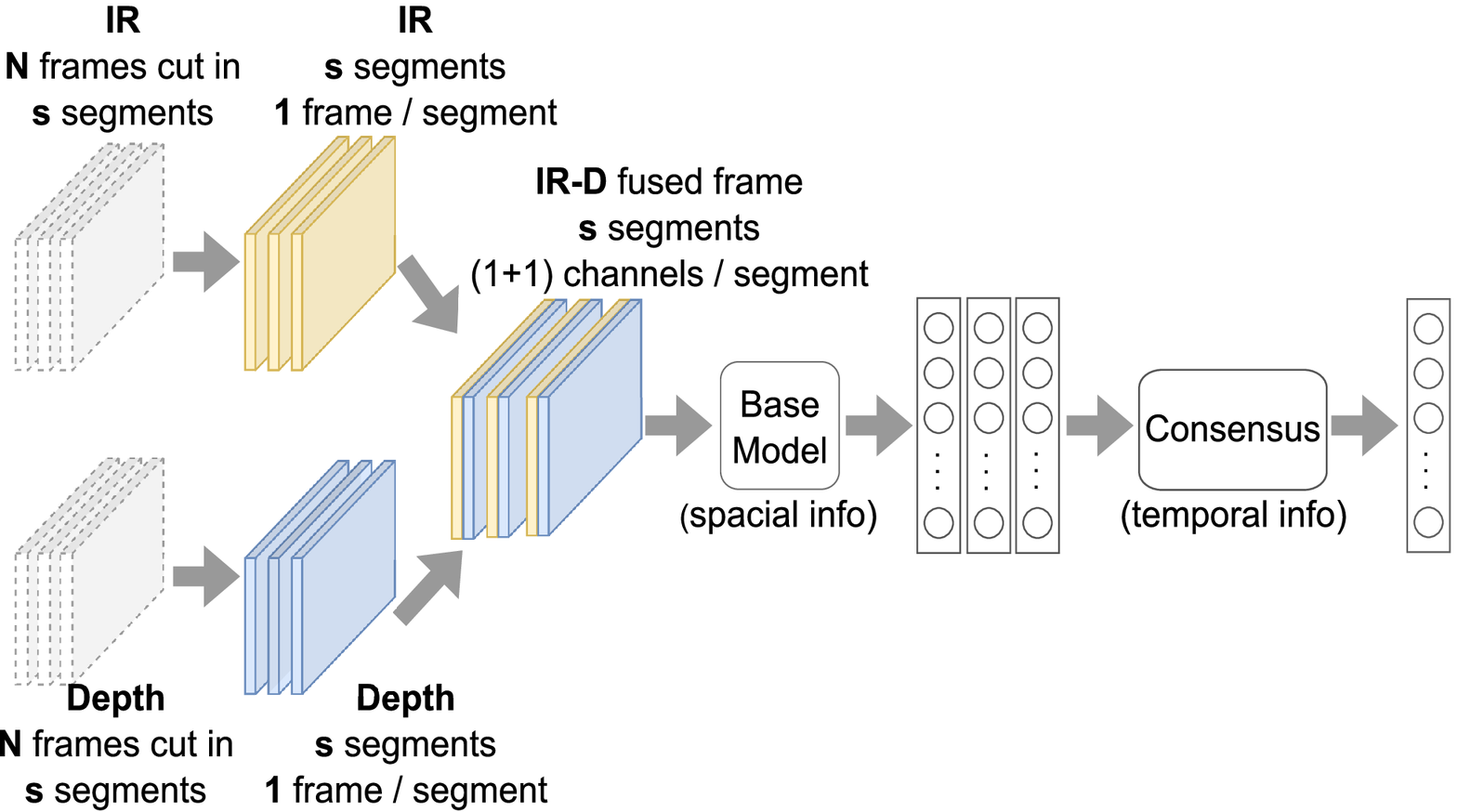}
        \caption{}
        \label{fig:flowchart_a}
    \end{subfigure}%
    \begin{subfigure}[t]{0.45\linewidth}
        \centering
        \includegraphics[height=3.3cm]{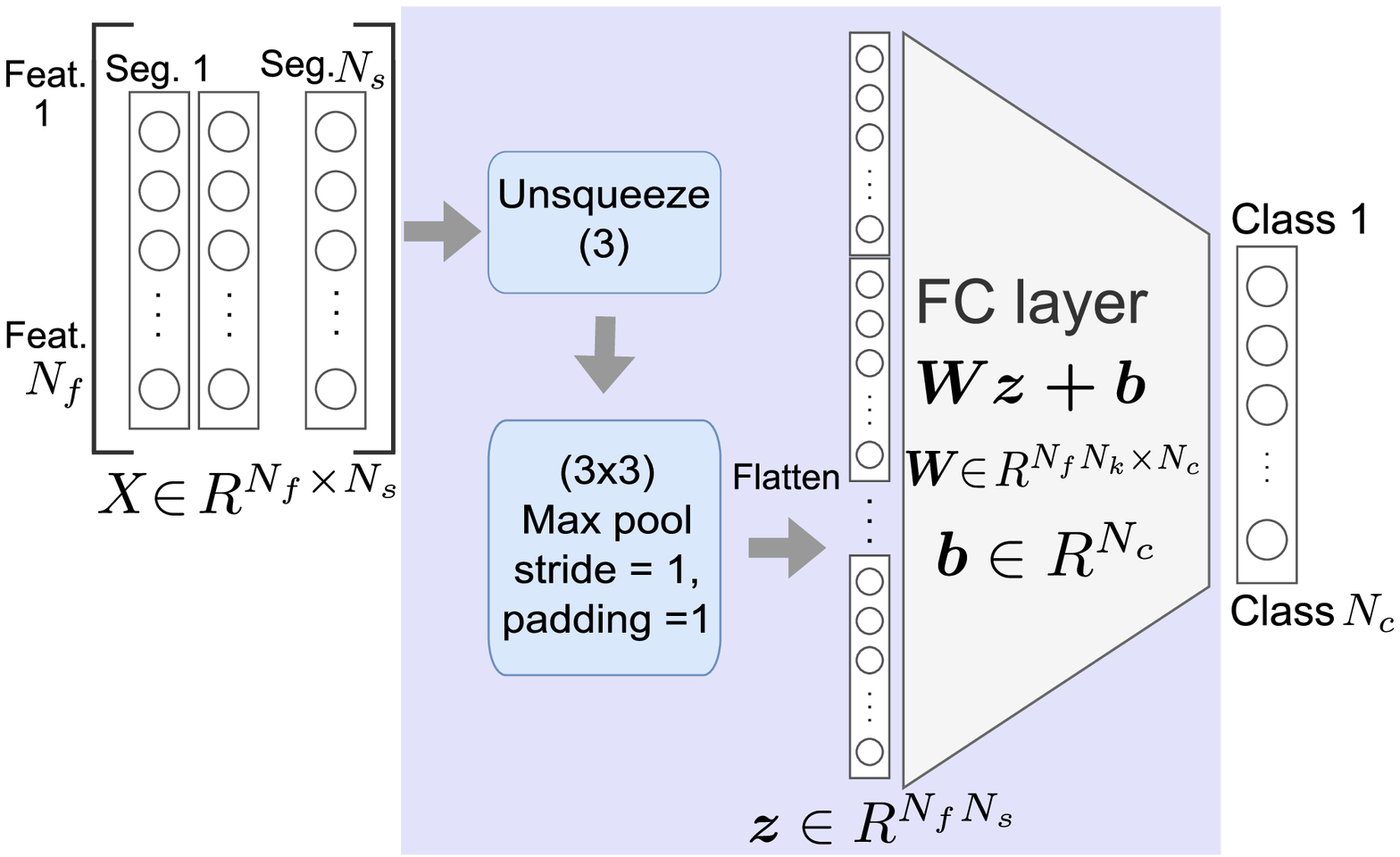}
        \caption{}
        \label{fig:flowchart_b}
    \end{subfigure}\\[1ex]
    \begin{subfigure}[t]{\linewidth}
        \centering
        \includegraphics[height=2.5cm]{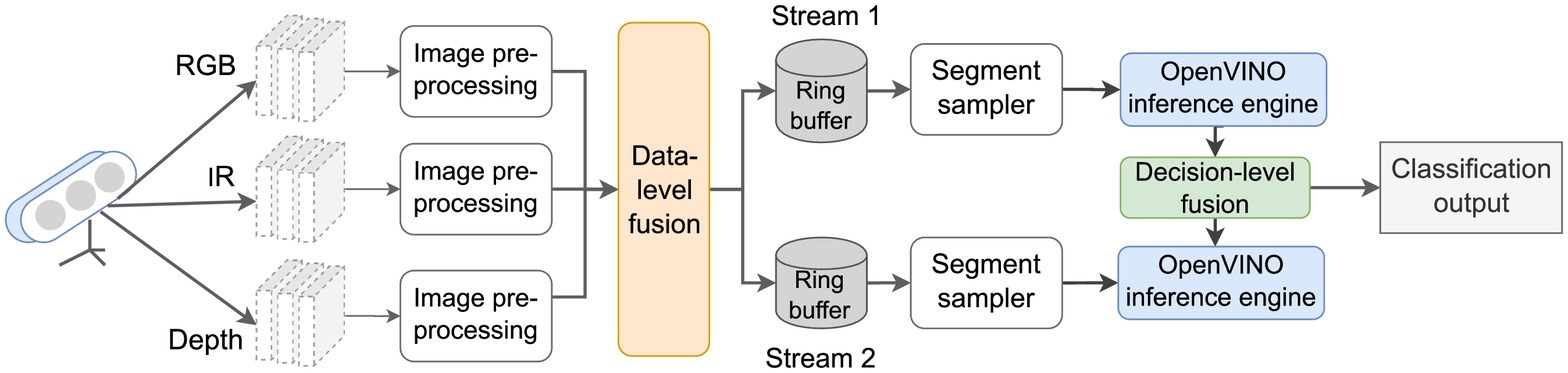}
        \caption{}
        \label{fig:inference}
    \end{subfigure}
	\caption{Proposed approach. (a) Training process flowchart (with fusion of RGB, IR and Depth). (b) Consensus module to capture temporal correlation (c) Real-time inference pipeline using the optimised OpenVINO models as engine.}
	\label{fig:flowchart}
\end{figure}

This work targets a real-world application hence our aim is to find the optimum architecture by analysing the use of different types of architectures for driver monitoring considering all high classification accuracy, fast reaction time, resource efficiency and architectures that can run on cost-efficient CPU-only platforms. 

The architecture is built on top of the TSN~\cite{wang16_eccv} work, using its partial batchnorm, and
data augmentation methods, while disabling horizontal flipping for the dBehaviourMD dataset. The batch size is chosen as 32 or 64 depending on the number of segments selected for the evaluations with 2D-CNN models. The stochastic gradient descent (SGD) is used as the optimiser with initial learning rate of 0.001, momentum 0.9 and weight decay $5 \times 10^{-4}$. The dropout rate before the final layer is set as 0.5. The learning rate is reduced by the learning rate decay factor (0.1) at each of the learning rate decay steps ([15, 30]). Training is finished after a maximum number of 40 epochs.

As shown in Fig. \ref{fig:flowchart}, modality fusion can be performed either at data level, decision level or both levels. Inspired by the data-level fusion of RGB and optical flow in MFF \cite{kopu18_cvprw}, in this paper, the fusion of RGB, IR, and depth is evaluated.

We have also explored different consensus modules. We apply max pooling for temporal dimension using the extracted features from each segment, which has the size of $X\in R^{N_{f}N_{s}}$ where $N_f$ and $N_s$ represent number of features and number of segments, respectively (Fig. \ref{fig:flowchart} (b)). The output is flattened and then passes through an multi-layer perceptron (MLP) layer. Experimental analysis shows that our consensus module outperforms the existing consensus modules such as averaging and MLPs applied in \cite{wang16_eccv} and \cite{kopu18_cvprw}, respectively.

In addition to the deterministic decision-level fusion such as taking average of the confidence scores of each classifier with different modality, this study also explores methods with prior knowledge of how each classifier performs on the validation set, namely the Bayesian weighting and Dempster–Shafer theory (DST) \cite{xu92_tsmc,sari07_tkde}.

\section{Experiments and results}
\label{sec:results}
In this section, we initially evaluate the recognition performances with both 2D-CNN models and 3D-CNNs using different types of feature extractors as base model in order to compare 2D and 3D CNNs and also to understand the impact of base model on recognition performances. 

Next, we apply multi-modal fusion in different levels such as data level and decision level. Finally, we provide runtime performances on CPU-only platform with the use of OpenVino for different modalities.

\subsection{Comparison of 2D CNN based models and 3D CNNs}

In order to analyse the impact of different architectures (i.e. base models in Fig.~\ref{fig:flowchart}) on recognition performances, with each of RGB, IR and depth modalities, we selected one dense and two resource efficient architectures for both 2D and 3D CNN based models, which are \textit{Inception-v3} \cite{szeg16_cvpr,szeg17_aaai} and \textit{3D-ResNeXt} \cite{hara18_cvpr} as dense architectures and 2D and 3D versions of \textit{MobileNet-v2} \cite{sand18_cvpr,kopu19}  and \textit{ShuffleNet-v2} \cite{ma18_eccv,kopu19} as resource efficient architectures for our evaluations in Table~\ref{tab:exp_base_mod}.

ResNet \cite{he16_cvpr} introduces residual learning and skip paths to achieve deeper architecture. Further improvements push CNN-based models to achieve the accuracy near or beyond human ability such as Inception series 
with inception modules which are ensembles of kernels of various sizes and bottleneck layers. Apart from targeting high classification accuracy, several works propose cost-effective models which can achieve slightly lower accuracy with greatly reduced computational effort. MobileNet-v2 and ShuffleNet-v2 are the two series of widely applied models in mobile platforms for real-time performance.

In this work, we also evaluated the impact of the dataset used for transfer learning using the same models for both 2D and 3D CNN based analysis (e.g. MobileNet-v2 and 3d-MobileNet-v2 for 2D and 3D based analysis, respectively) for fair comparison. Here, we used ImageNet dataset and Kinetics-600 dataset for the transfer learning of 2D and corresponding 3D CNN models, respectively.

For the training of 3D-CNNs, SGD with standard categorical cross entropy loss is applied and largest
fitting batch size is selected for each CNN model. The momentum, dampening and weight decay are set to 0.9, 0.9 and $1\times10^{-3}$, respectively. For the training of
dBehaviourMD, we have used the pre-trained models
on Kinetics-600 provided by \cite{kopu19}. For fine-tuning, we start
with a learning rate of 0.1 and reduced it two times after
$20^{th}$ and $35^{th}$ epochs with a factor of $10^{-1}$. Optimisation is completed after 10 more epochs. A
dropout layer is applied before the final layer of
the networks with a ratio of 0.5.

For temporal augmentation, input clips are selected from a random temporal position in the video
clip. For the videos containing smaller number of frames than
the input size, loop padding is applied. Since the average number of frames in our dataset is around 50 frames and the input to
the networks is selected as 16-frame clips, in order to capture the most content of the action in the input clip, we have
applied downsampling of 2, which enables to capture temporal information for 32-frame extent. Similar types of spatial augmentation is applied for both 2D and 3D CNN based models. The spatial resolution of the input is selected as 224 and 112 for 2D and 3D CNN networks, respectively.

\begin{table}[tbp]
    \caption{Behaviour recognition performances with 2D and 3D CNN based models.}
    \label{tab:exp_base_mod}
    \centering
    \begin{tabular}{lllcc}
      \hline
      {Modality} & {Architecture} & {Base Model} & {Segments} & {Top-1 Acc. (\%)} \\
       &  &  &  &  \\
      \hline
      \hline
      RGB & 2D CNN & Inception-v3   & 4/8 & 93.2/91.8  \\ 
      & & MobileNet-v2 1.0x & 4/8 & 88.7/89.9  \\ 
      & & ShuffleNet-v2 1.0x   & 4/8 & 81.4/84.8  \\
      \cline{2-5}
       & 3D CNN & 3D-ResNeXt  & - & 91.6  \\
       &  & 3D-MobileNet-v2 1.0x  & - & 90.3  \\
       &  & 3D-ShuffleNet-v2 1.0x  & - & 88.5  \\       
      \hline \hline
      IR & 2D CNN & Inception-v3    & 4/8 & 92/91.6  \\
      & & MobileNet-v2 1.0x     & 4/8 & 90.1/91.1  \\
      & & ShuffleNet-v2 1.0x  & 4/8 & 82.6/85.4  \\ 
      \cline{2-5}
       & 3D CNN & 3D-ResNeXt  & - & 92.9  \\
       &  & 3D-MobileNet-v2 1.0x  & - & 92.7  \\
       &  & 3D-ShuffleNet-v2 1.0x  & - & 90.6  \\    
      \hline \hline
      Depth & 2D CNN & Inception-v3   & 4/8 & 90.1/91.9  \\
      & & MobileNet-v2 1.0x   & 4/8 & 91.6/91.7  \\
      & & ShuffleNet-v2 1.0x  & 4/8 & 83.4/88.1  \\
      \cline{2-5}
       & 3D CNN & 3D-ResNeXt  & - & 92.8  \\
       &  & 3D-MobileNet-v2 1.0x  & - & 92.4  \\
       &  & 3D-ShuffleNet-v2 1.0x  & - & 89  \\    
      \hline 
    \end{tabular}
\end{table}

Table \ref{tab:exp_base_mod} shows our results with the approach in Fig.~\ref{fig:flowchart} using averaging as consensus module, which simply takes the average of the scores computed for each selected frame of an action video. According to the results in Table \ref{tab:exp_base_mod}:
\begin{itemize}
\item The overall accuracy can approach around 90\%, even with light-weight feature extractor as MobileNet-v2 for RGB modality. 
\item The benefit of using high-performance base models is more obvious for RGB.
\item Depth modality performs better than RGB and IR modality. For the depth and IR modality,
light-weight base model MobileneNet-v2 is enough to reach its accuracy limit.
\item The number of segments $N_s=4$ already reaches the accuracy bound for many models compared to $N_s=8$ case.
\item 2D CNN based models can reach almost the similar accuracy of their 3D counterparts for MobileNet-v2 and dense architectures. This result also shows that 2D models pre-trained on a large-scale image dataset (ImageNet dataset) performs as good as 3D models pre-trained on video dataset (Kinetics-600 dataset) as stated in \cite{hara18_cvpr} as well.
\item IR performs slightly better than RGB. Since the dataset was recorded during the daytime, the major difference between RGB and IR is colour information. This might be the reason for IR models to easier learn the patterns such as shapes and poses compared to RGB models.
\end{itemize}

\subsection{Impact of different consensus modules on performance}

For temporal understanding of action videos, we have used several consensus modules and analysed their impact on recognition performances. Table \ref{tab:exp_base_vidas} shows the results we obtained with averaging and MLP based consensus modules in addition to our proposed consensus module (MaxP-MLP in Table \ref{tab:exp_base_vidas}).\footnote{$N_{f}$ is set to 11 for averaging-based consensus, excluding \textit{hands free} and \textit{switch gear} actions. $N_{f}$ is set to 64 for both MLP and MaxP-MLP in all experiments.} With the MobileNet-v2 as base model, the results show that the MaxP-MLP outperforms other consensus modules almost in all cases. The applied pooling operation in MaxP-MLP increases the accuracy considerably compared to other consensus modules since this process enables to emphasise the significant features in the spatio-temporal features group ($X$ in Fig. \ref{fig:flowchart_b}) representing an action video.

The evaluations are done on both validation and test sets for both 4 and 8 segment analysis. $N_s=4$ would already reach its performance upper bound. Similar to the previous outcome, depth modality provides better accuracy compared to RGB and IR. Data level fusion of IR and Depth modalities increases the accuracy on both validation and test sets with both 4 and 8 segment analysis.  

\begin{table}[t!]
\caption{Comparison of the proposed consensus module with averaging and MLP based consensus modules on both validation and test sets.}
\label{tab:exp_base_vidas}
\centering
\begin{tabular}{llllcccc|cccc}
  \hline
  {Base Model} & {Consensus}& {$N_{f}$} & {Seg.} & \multicolumn{8}{c}{Top-1 Acc. (\%)}  \\
    & & & & \multicolumn{4}{c}{Validation Set} & \multicolumn{4}{c}{Test Set} \\
   \cline{5-8} \cline{9-12}
    & & & & RGB & Depth & IR & IRD & RGB & Depth & IR & IRD \\
  \hline
  \hline
  {MobileNet-V2} & Average & 11 & 4 & 89.6 & 92.0 & 91.1 & - & 89.4 & 88.7 & 89.4 & - \\
  & MLP & 64 & 4 & 90.7 & 91.4 & 90.2 & - & 91.9 & 90.6 & 90.6 & -\\
  & MaxP-MLP & 64 &4 & \textbf{91.5} & \textbf{93.5} & \textbf{92.6} & \textbf{94.3} & \textbf{92} & \textbf{91.6} & \textbf{91.5} & \textbf{91.9} \\
  \cline{2-12}
  & Average & 11 & 8 & 90.4 & 93 & 91.6 & - & 88.9 & 92.3 & 89.5 & -\\
  & MLP & 64 & 8 & 91.2 & 93 & 91.1 & - & 91.0 & \textbf{93.2} & 91.6 & - \\
  & MaxP-MLP & 64 & 8 & \textbf{92.0} & \textbf{94.1} & \textbf{92.6} & \textbf{95} & \textbf{91.6} & 92.7 & \textbf{91.8} & \textbf{93.9} \\
  \hline
\end{tabular}
\end{table}

\subsection{Multi-modal fusion}
\label{ssec:model_fs}
  
Fusion of CNN based classifiers can be done at (i) data-level: two types of data are fused before fed into the CNN model \cite{kopu18_cvprw}; and at (ii) decision-level: results from two classifiers are fused together.

Table \ref{tab:fusion} shows the results for the applied data level and decision level fusion. Since IR and depth modalities have pixel-wise correspondence, these modalities are fused at data level which provides an enhanced accuracy compared to their mono-modal analysis. Applying decision level fusion for RGB (Stream 1) and IRD (Stream 2) modalities using the deterministic approach of averaging and the statistics methods based on the prior knowledge of the classifiers such as bayesian weighting and DST, we observed considerable increase in the performances compared to the results of single streams. However, the results also show that statistics methods do not help more for our evaluations.  

\begin{table}[tbp]
	\caption{Fusion of modalities in different levels. Stream1 (IRD) + Stream2 (RGB). Base model is MobileNet-v2. MaxP-MLP is the consensus module.}
	\label{tab:fusion}
	\centering
	\begin{tabular}{lcccc}
		\hline
		Fusion Level   & Fusion Type & Modalities & Segments & Top-1 Acc (\%) \\
		\hline
		Data level & - & IRD & 4 & 91.9 \\
		\hline
		- & - & RGB & 4 &  92  \\
		\hline
		Decision level &  Averaging fusion & RGB + IRD & 4 & 93.7  \\
		 &  Bayesian weighting & RGB + IRD & 4 &  93.7  \\
		& DST & RGB + IRD & 4 & 93.7  \\
		\hline
	\end{tabular}
\end{table}

\subsection{Runtime performance analysis}

The proposed architecture is implemented as a real-time driver behaviour monitoring system using OpenVINO (version 2020.4.255) \cite{opvn} and selecting MobileNet-v2 as base model for 4-segment analysis. Our results show that MobileNet-v2 provides comparable accuracy with dense architectures with the advantage of having much less number of trainable parameters, achieving real-time deployment, high accuracy, resource efficiency and fast reaction time. The selected models are converted to OpenVINO IR and tested with the inference pipeline on an Intel Core-i9 7940X machine. The single modalities RGB, IR and depth can run
up to 8.1 \textit{ms}, 7.3 \textit{ms} and 7.3 \textit{ms} per action video, respectively. The model which is based on data level fusion of IR and Depth modalities can run up to 10.1 \textit{ms} and two streams (RGB and IRD) together can still reach 18 \textit{ms}, enabling the whole
inference pipeline to operate at more than 30 \textit{FPS}.

\section{Conclusions}
\label{sec:conclusions}

In this paper we present the Driver Monitoring Dataset (DMD) as an extensive, varied and comprehensive dataset created for development of Driver Monitoring Systems (DMS), specially designed for training and validating DMS in the context of automated driving SAE L2-3. The dataset has been devised to be multi-purpose, including footage of real driving scenes, with scenes containing driving actions related to driver's distraction, fatigue and behaviour. The dataset is multi-modal as it contains 3 cameras and 3 streams per camera (RGB, Depth, IR), and labels covering a wide range of visual features (e.g. time-lapse actions, body and head-pose, blinking patterns, hands-wheel interactions, etc.).

This paper also reports our work using DMD to extract the \textit{dBehaviourMD} training set, crafted to train a DL model proposed for a real-world application for real-time driver behaviour recognition. The achieved performance of the trained model yields enhanced accuracy for different multi-modal fusion approaches while keeping real-time operation on CPU-only platforms.

\vspace{-5mm}
\subsubsection*{Acknowledgements} 
We would like to thank all the people who participated in the recording process for sharing their time to create the DMD dataset.
This work has received funding from the European Union's H2020 research and innovation programme (grant agreement no 690772, project VI-DAS).

\clearpage


\end{document}